\documentclass[final,5p,times,twocolumn]{elsarticle}
\usepackage{bm}  
\usepackage{flushend}

\usepackage{amssymb}
\usepackage{amsmath}

\journal{Nuclear Physics B}

\begin{document}
	\graphicspath{{Pictures/}}
\begin{frontmatter}



\title{Trajectory Tracking and Stabilization of Quadrotors Using\\ Deep Koopman Model Predictive Control}


\author{Haitham El-Hussieny} 

\affiliation{organization={Department of Mechatronics and Robotics Engineering, Egypt-Japan University of Science and Technology (E-JUST)},
            city={\\New Burg El-Arab},
            postcode={21934}, 
            state={Alexandria},
            country={Egypt}}

\begin{abstract}
	This paper presents a data-driven control framework for quadrotor systems that integrates a deep Koopman operator with model predictive control (DK-MPC). The deep Koopman operator is trained on sampled flight data to construct a high-dimensional latent representation in which the nonlinear quadrotor dynamics are approximated by linear models. This linearization enables the application of MPC to efficiently optimize control actions over a finite prediction horizon, ensuring accurate trajectory tracking and stabilization. The proposed DK-MPC approach is validated through a series of trajectory-following and point-stabilization numerical experiments, where it demonstrates superior tracking accuracy and significantly lower computation time compared to conventional nonlinear MPC. These results highlight the potential of Koopman-based learning methods to handle complex quadrotor dynamics while meeting the real-time requirements of embedded flight control. Future work will focus on extending the framework to more agile flight scenarios and improving robustness against external disturbances.
\end{abstract}




\end{frontmatter}



\section{Introduction}
\label{sec:introduction}
Quadrotor aerial robots, with their lightweight design and agile six-degree-of-freedom motion, have emerged as highly versatile platforms for research and practical applications~\cite{foehn2022agilicious, zulu2014review}. Their ability to perform rapid maneuvers, hover with precision, and operate in constrained environments makes them suitable for diverse tasks such as aerial surveillance~\cite{aizelman2024quadrotor}, package delivery~\cite{saunders2024autonomous}, environmental monitoring~\cite{kokate2023review}, and search-and-rescue missions~\cite{lyu2023unmanned}. Furthermore, quadrotors are widely used as testbeds for advanced control strategies and embodied intelligence~\cite{khalid2023control}. However, achieving reliable and high-performance control remains a challenge due to the strongly nonlinear dynamics of quadrotor flight, the tight coupling between translational and rotational states, and the sensitivity to disturbances such as wind gusts and payload variations. These factors make precise modeling and control design difficult when relying solely on traditional analytical approaches~\cite{dhadekar2021robust,hanover2021performance,wang2016trajectory}.

To tackle these challenges, two dominant control methodologies have emerged in quadrotor research: model-based control and learning-based control~\cite{sonmez2024survey}. Model-based strategies often employ simplified dynamics, such as linearization around hover conditions using LQR or feedback linearization techniques, to effectively approximate the quadrotor's behavior under specific assumptions~\cite{belkheiri2012different, khalid2023control}. While these approaches offer structured frameworks for control design, their reliance on simplifying assumptions can lead to model-plant mismatches and suboptimal performance when faced with real-world disturbances or agile maneuvers~\cite{salzmann2023real}.

To overcome the limitations of traditional model-based control, modern research has increasingly turned to learning-based strategies that leverage deep learning methods to better capture complex quadrotor dynamics and couplings~\cite{saviolo2022physics}. One promising approach involves learning inverse dynamics: neural networks trained to map desired trajectories directly to actuator commands, simplifying controller implementation and enabling agile flight in challenging scenarios~\cite{zhou2018inversion, el2024real}. However, the inherently underactuated nature of quadrotors and the possibility of multiple valid inverse mappings introduce challenges for these methods, potentially affecting generalization and stability.

Another avenue explored is reinforcement learning (RL) \cite{el2024obstacle}, where control policies are trained in simulation, either model-free or with learned models, and then deployed on real quadrotors~\cite{hwangbo2017control}. While RL offers the flexibility to discover effective control strategies through experience, its deployment is often hindered by the sim-to-real gap. Transfers from simulation to real-world quadrotors can falter unless substantial domain randomization, real-world adaptation, or robust policy learning techniques are applied~\cite{dionigi2024benchmarking}.

To overcome the difficulties of modeling forward dynamics in quadrotor systems, Koopman operator–based methods have gained momentum, offering a compelling strategy for control design~\cite{narayanan2023se, abido2024koopman, martini2024koopman}. By embedding the nonlinear dynamics into a lifted space where the system evolves linearly, these approaches enable the deployment of efficient linear control techniques. However, a critical challenge lies in defining appropriate lifting (observable) functions, insufficient or suboptimal choices can introduce substantial modeling errors, negatively impacting prediction fidelity and controller effectiveness~\cite{rajkumar2024analytical, shi2024lifting}.

In this work, we introduce a deep Koopman-based model predictive control (DK-MPC) framework for quadrotor systems, extending prior Koopman-based approaches that have largely been restricted to rigid-body platforms or simulation studies. Our method employs deep neural networks (DNNs) to automatically learn suitable embeddings, thereby improving the predictive capability of the Koopman operator. This results in a globally linearized representation of the quadrotor dynamics, which serves as the dynamic constraint within the MPC formulation to enable accurate trajectory tracking. The integration of the learned Koopman model with MPC enables real-time optimization of control inputs, ensuring high-precision and computationally efficient control performance on physical quadrotor platforms.

Numerical simulation experiments on a data-driven quadrotor model validate the effectiveness of the proposed DK-MPC framework. The results show that DK-MPC achieves highly accurate control performance, with clear improvements over conventional nonlinear control strategies. These outcomes highlight the potential of DK-MPC for future quadrotor applications, particularly in scenarios requiring precise trajectory tracking and robustness against dynamic variations~\cite{narayanan2023se}. Moreover, as a data-driven method, DK-MPC can be easily adapted to different quadrotor platforms, providing a flexible and scalable solution for advanced aerial robotics.

The remainder of this paper is organized as follows: Section~\ref{sec:method} presents the development of the proposed DK-MPC framework, highlighting the integration of the deep Koopman operator with model predictive control. Section~\ref{sec:expr} reports the results of numerical simulation experiments, demonstrating the effectiveness of DK-MPC in enabling the quadrotor to accurately track reference trajectories while respecting system constraints. Finally, Section~\ref{sec:conclude} concludes the paper with a summary of key findings and outlines directions for future research.

\section{Deep Koopman-based Model Predictive Control}
\label{sec:method}
This section presents a comprehensive overview of the methodology behind the proposed Deep Koopman-based Model Predictive Control (DK-MPC) approach, tailored to tackle the control complexities of quadrotors. It begins with an introduction to the overall architecture, followed by an explanation of the Koopman operator's formulation and learning process in high-dimensional settings, and concludes with the integration of the learned Koopman model into the MPC framework for control implementation.

\subsection{Overview of the DK-MPC Framework}

\begin{figure}[!p!t]
	\centering
	\includegraphics[width=.8\columnwidth]{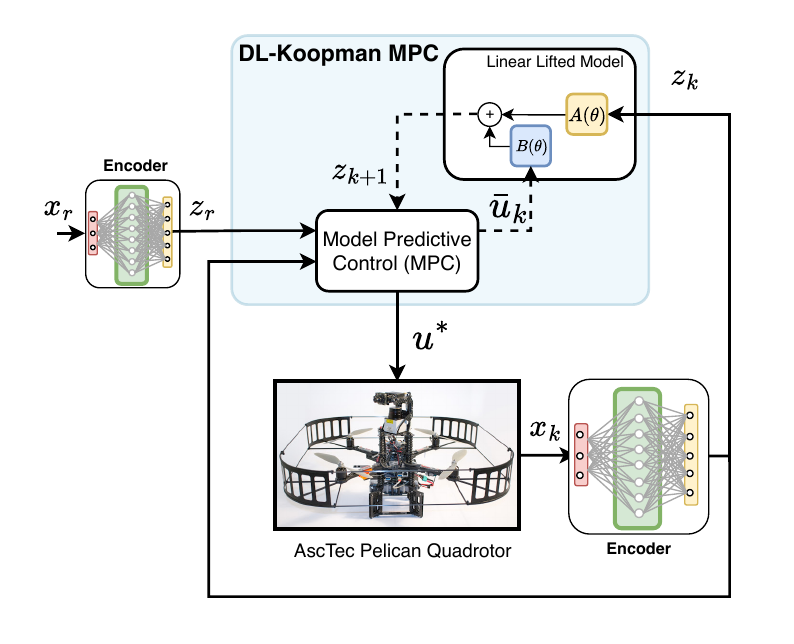}
	\caption{Schematic of the proposed DK-MPC framework for controlling the AscTec Pelcian Quadrotor. The deep Koopman operator projects both the reference state $\bm{\mathit{x}}_{ref}$ and the current state $\bm{\mathit{x}}$ into a high-dimensional latent space where the system exhibits linear behavior. Using these latent representations and the corresponding linear dynamics, the MPC controller computes optimal control inputs 	$\bm{\mathit{u}}^*$ to guide the quadrotor along the desired trajectory.}\label{fig:mpc_dk}
\end{figure}

Figure \ref{fig:mpc_dk} depicts the structure of the proposed DK-MPC approach. It utilizes a deep learning-based Koopman operator to create a globally linear, time-invariant representation of the nonlinear system by mapping the original state space into a higher-dimensional latent space. During each control cycle, the deep Koopman encoder maps the reference state $\bm{\mathit{x}}_{ref}$ into its corresponding high-dimensional latent representation. Simultaneously, the current state $\bm{\mathit{x}}$, obtained from the quadrotor, is encoded into the same latent space. Within this transformed space, an MPC algorithm optimizes a quadratic cost function based on the latent states $\bm{\mathit{z}}$ and $\bm{\mathit{z}}_{ref}$ over a finite prediction horizon, thereby producing the optimal control input $\bm{\mathit{u}}^*$ to follow the reference trajectory.

\subsection{Koopman Operator for Linear Representation of Quadrotor Dynamics}
We examine a discrete-time nonlinear system governed by a function $f$, which describes the evolution of the state in the quadrotor. In this context, $\bm{\mathit{x}}_k$ denotes the system’s state at time step $k$, and $\bm{\mathit{u}}_k$ represents the corresponding control input, such that:

\begin{equation}
	\label{eq: system_model}
	\bm{\mathit{x}}_{k+1} =  f(\bm{\mathit{x}}_k, \bm{\mathit{u}}_k )
\end{equation}
Due to the inherent complexity and nonlinearity of $f$, directly designing a control strategy can be challenging. To address this, we apply the Koopman operator framework, which operates in a transformed (lifted) space defined by a mapping function $\phi$. This function lifts the original nonlinear system into a higher-dimensional space where its behavior becomes approximately linear:

\begin{equation}
	\label{eq:koopman}
	\phi(f(\bm{\mathit{x}}_k, \bm{\mathit{u}}_k),\bm{\mathit{u}}_{k+1})\,=\ K\phi(\bm{\mathit{x}}_k, \bm{\mathit{u}}_k)
\end{equation}

The effectiveness of this linear approximation relies heavily on the choice of the lifting function 
$\phi$. In this study, we adopt a deep learning-based method to learn $\phi$ enabling us to model the complex nonlinear dynamics typical of quadrotors. 

Extending the Koopman operator framework, the lifting function $\phi(x, u)$ is separated into two components: one dependent on the state, $\phi_x(x)$, and the other on the input, $\phi_u(u)$, such that:
\begin{equation}
	\label{eq:phi}
	\phi(x, u) = 
	\begin{bmatrix}
		\phi_x(x) \\
		\phi_u(u)
	\end{bmatrix}
\end{equation}

This decomposition enables the system dynamics to be expressed in a control-affine form through matrix operations:
\begin{equation}
	\label{eq:phi2}
\begin{bmatrix}
	\phi_x(x_{k+1}) \\
	\phi_u(u_{k+1})
\end{bmatrix}
=
\begin{bmatrix}
	K_{xx} & K_{xu} \\
	K_{ux} & K_{uu}
\end{bmatrix}
\begin{bmatrix}
	\phi_x(x_k) \\
	\phi_u(u_k)
\end{bmatrix}
\end{equation}

From this, the evolution of the lifted state can be written as:
\begin{equation}
	\label{eq:phi3}
\phi_x(x_{k+1}) = K_{xx} \phi_x(x_k) + K_{xu} \phi_u(u_k)
\end{equation}

Consistent with simplifications proposed in earlier studies~\cite{bruder2020data, shi2022deep}, the input lifting function $\phi_u(u)$ is directly represented by the control input $u$. Defining $K_{xx} = A$, $K_{xu} = B$, and $\phi_x(x_k) = z_k$, the dynamics simplify to a globally linear form:

\begin{equation}
	\label{eq:linear}
\bm{\mathit{z}}_{k+1} = A \bm{\mathit{z}}_k + B \bm{\mathit{u}}_k
\end{equation}

This transformation yields a linear state-space model that not only facilitates controller synthesis but also effectively approximates the overall behavior of the original nonlinear quadrotor system.

\subsection{Deep Learning Approach for Approximating the Koopman Operator}
\begin{figure*}[!p!t]
	\centering
	\includegraphics[width=.8\textwidth]{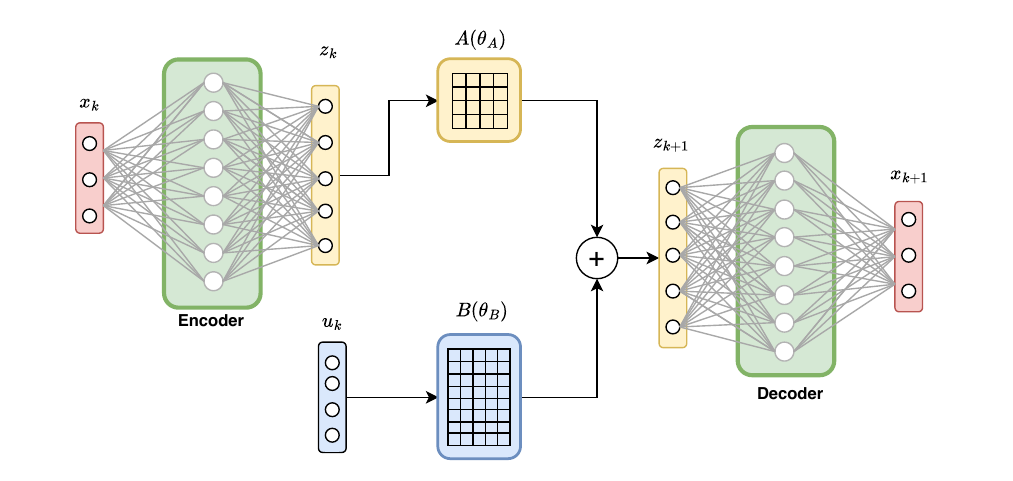}
	\caption{Architecture of the proposed Deep Koopman-based framework. The encoder maps the quadrotor state $x_k$ into a latent space representation $z_k$. The dynamics in this latent space are modeled linearly using matrices $A(\theta_A)$ and $B(\theta_B)$, which evolve the system according to the control input $u_k$. The decoder then reconstructs the next state $x_{k+1}$ from the predicted latent state $z_{k+1}$. This integration of neural networks with Koopman-based linear dynamics enables accurate forward prediction and serves as the foundation for the DK-MPC controller.}\label{fig:framework}
\end{figure*}
This subsection outlines the deep learning strategy employed to estimate the Koopman operator and its associated lifting and inverse functions. As depicted in Fig.~\ref{fig:framework}, the method is based on a deep auto-encoder architecture designed to learn the forward mapping $\phi$ and its inverse $\phi^{-1}$. The auto-encoder comprises two primary components: an encoder and a decoder. The encoder $\phi$, implemented using a Multi-Layer Perceptron (MLP), lifts the system's original state into a higher-dimensional latent space, where the system dynamics become approximately linear. Conversely, the decoder $\phi^{-1}$ reconstructs the original state from this latent representation.

To model the system's linear behavior in the latent space, two additional single-layer MLPs are employed to approximate the matrices $A$ and $B$ from Eq.~\eqref{eq:linear}. These networks are designed without biases or activation functions, enabling them to function as straightforward linear transformations. This structure ensures that the learned latent dynamics remain consistent with the theoretical assumptions of the Koopman framework.

\textbf{Reconstruction Loss:} To accurately model the relationship between the original and latent state representations, a reconstruction loss is defined using the $L_2$ norm. This loss ensures that the auto-encoder learns to map states to and from the latent space with minimal error:
\begin{equation}
	L_{\text{recon}} = \left\| \bm{\mathit{x}}_k - \phi^{-1}(\phi(\bm{\mathit{x}}_k)) \right\|_2^2,
	\label{eq:recon_loss}
\end{equation}

\textbf{Linear Dynamics Loss:} To identify the linear operator $A$ and the control matrix $B$, we define a dynamics loss that encourages the latent state transitions to follow a linear model structure. This objective penalizes deviations from the expected linear behavior in the latent space:
\begin{equation}
	L_{\text{linear}} = \left\| \phi(\bm{\mathit{x}}_{k+1}) - (A\phi(\bm{\mathit{x}}_k) + B \bm{\mathit{u}}_k) \right\|_2^2,
	\label{eq:linear_loss}
\end{equation}

\textbf{Stability Loss:} To promote stability in the learned Koopman operator $A$, we introduce a stability loss that penalizes cases where the spectral radius exceeds $1$. Specifically, we compute the spectral radius $\rho(A)$ as the maximum absolute eigenvalue of $A$, and apply a ReLU-based penalty to constrain it within the unit circle:
\[
\rho(A) = \max \left( \left| \lambda_i \right| \right), \quad \lambda_i \in \text{eig}(A)
\]

\begin{equation}
L_{\text{stability}} = \left( \max(0, \rho(A) - 1) \right)^2
	\label{eq:stability_loss}
\end{equation}

This loss term helps ensure the linear dynamics model remains stable by discouraging eigenvalues with magnitudes greater than one.

\textbf{Total Loss Function:} The final objective, as shown in Eq.~\eqref{eq:total_loss}, is expressed as a weighted sum of the reconstruction loss, linear dynamics loss, stability loss, along with an $L_2$ regularization term to reduce overfitting~\cite{ng2004feature}:
\begin{equation}
	L = \lambda_1 L_{\text{recon}} + \lambda_2 L_{\text{linear}} + \lambda_3 L_{\text{stability}} + \lambda_4 \| W \|_2^2
	\label{eq:total_loss}
\end{equation}

This integrated framework supports end-to-end learning of the Koopman-based model by jointly training the linear operator $A$, the control-affine matrix $B$, and the lifting function $\phi$. Consequently, it enables effective modeling and control of the nonlinear dynamics inherent in quadrotor systems.

\subsection{Integrating MPC with the Deep Koopman Operator}

Once the Koopman operator has been learned, we employ Model Predictive Control (MPC) to regulate the nonlinear behavior of the quadrotor. MPC is particularly well-suited for systems with constraints and excels in optimizing control trajectories by predicting future system evolution~\cite{el2024real}. Utilizing the linear system representation obtained via the Koopman framework, we can formulate an MPC scheme that is both efficient and effective.

The controller solves an optimization problem to compute an optimal sequence of control actions $\hat{\bm{\mathit{u}}}_{t:t+H}^*$ over a finite prediction horizon $H \in \mathbb{N}$. The cost function penalizes both the deviation from the reference trajectory in the lifted space and the magnitude of the control inputs:

\begin{equation}
	\label{eq:mpc}
	\begin{aligned}
		\min_{\hat{\bm{\mathit{u}}}_{t:t+H}} \quad & \sum_{k=0}^{H} 
		\left( 
		\hat{\bm{\mathit{z}}}_{t+k} - \hat{\bm{\mathit{z}}}^{\text{ref}}_{t+k}
		\right)^T 
		\hat{Q}
		\left( 
		\hat{\bm{\mathit{z}}}_{t+k} - \hat{\bm{\mathit{z}}}^{\text{ref}}_{t+k}
		\right)
		+ 
		\hat{\bm{\mathit{u}}}_{t+k}^T \hat{R} \hat{\bm{\mathit{u}}}_{t+k} \\
		\text{s.t.} \quad 
		& \hat{\bm{\mathit{z}}}_{t+k} = A \hat{\bm{\mathit{z}}}_{t+k-1} + B \hat{\bm{\mathit{u}}}_{t+k-1}, \quad k = 1, 2, \ldots, H \\
		& \hat{\bm{\mathit{z}}}_{t} = \phi(\bm{\mathit{x}}_t), \\
		& \hat{\bm{\mathit{z}}}^{\text{ref}}_{t+k} = \phi(\bm{\mathit{z}}^{\text{ref}}_{t+k}), \quad k = 1, 2, \ldots, H \\
		& \bm{\mathit{u}}_{\min} \leq \hat{\bm{\mathit{u}}}_{t+k} \leq \bm{\mathit{u}}_{\max}, \quad k = 0, 1, \ldots, H
	\end{aligned}
\end{equation}

Here, $\hat{Q} \in \mathbb{R}^{n \times n}$ and $\hat{R} \in \mathbb{R}^{m \times m}$ are positive semi-definite weighting matrices that penalize state tracking error and control effort, respectively. The variables $\bm{\mathit{u}}_{\min}$ and $\bm{\mathit{u}}_{\max}$ define the allowable range for the control inputs, while $n$ denotes the dimension of the latent (lifted) state space.
At each control timestep, we encode the current system state $\bm{\mathit{x}}_t$ and future reference states $\bm{\mathit{x}}^{\text{ref}}_{t:t+H}$ into the latent space, yielding $\hat{\bm{\mathit{z}}}_t$ and $\hat{\bm{\mathit{z}}}^{\text{ref}}_{t:t+H}$. Solving the optimization problem in Eq.~\eqref{eq:mpc} produces an optimal control sequence, from which only the first control action $\hat{\bm{\mathit{u}}}^*_t$ is applied to the system. This process is repeated iteratively until convergence is achieved.

\section{Experiments}
\label{sec:expr}
This section outlines the methodology for data acquisition, and the training procedure for the proposed model. We then deploy the DK-MPC controller to manage the motion of the quadrotor.  To validate the effectiveness of the proposed approach, we conduct three experimental scenarios: a path-following task, a moving target tracking task, and a dynamic obstacle avoidance task. The first experiment highlights the precision and fast dynamic response achieved by DK-MPC, the second illustrates its capability to adapt to dynamic conditions, whil the third assess the responsivness of the control system to handle avoidance of dynamic obstacles, indicating its potential for real-world deployment in responsive control applications.

\subsection{Platform Description: AscTec Pelican Quadrotor}
To evaluate the proposed modeling and control strategies in a dynamic aerial platform, we employ the AscTec Pelican quadrotor, a research-grade unmanned aerial vehicle (UAV) known for its modularity and robust performance. The quadrotor features a rigid carbon-fiber frame with a diagonal motor-to-motor span of approximately $85$\ cm and a total weight of around $1.6$\ kg, including battery and onboard electronics. It is powered by four brushless DC motors equipped with fixed-pitch propellers and is capable of both high-thrust maneuvers and stable hovering.

The Pelican is equipped with an onboard computer running a real-time operating system, supporting integration with ROS for high-level control and sensor fusion. It includes an IMU, barometer, and GPS module for autonomous navigation, and offers payload options for additional sensors or manipulators. The platform's flexibility and open software interface make it ideal for implementing and validating advanced control algorithms, including the proposed Koopman-based MPC framework.

\subsection{Dataset for DK-MPC Training}

For training and evaluating the proposed DK-MPC framework, we utilize the publicly available Pelican Dataset provided by the WaveLab group~\cite{mohajerin2018deep, mohajerin2019multistep}. This dataset was collected using the AscTec Pelican quadrotor in a controlled indoor environment equipped with a motion capture system. The dataset includes time-series recordings of the quadrotor's state and control inputs across various flight maneuvers.

Specifically, the dataset contains measurements such as position, velocity, acceleration, orientation (quaternions), angular rates, and motor commands, all sampled at 100\,Hz. These high-fidelity recordings offer a comprehensive representation of the quadrotor’s dynamic behavior and are well-suited for learning both the lifting functions and the Koopman operator used in the DK-MPC framework.

To ensure robust training, we preprocess the dataset by segmenting it into input-output pairs $(\bm{\mathit{x}}_k, \bm{\mathit{u}}_k, \bm{\mathit{x}}_{k+1})$, normalizing each feature, and partitioning the data into training and validation sets. The richness and quality of this dataset make it ideal for modeling complex nonlinear dynamics in aerial robotics.

\subsection{Model Training}

To facilitate robust learning, the dataset was divided into three subsets: training, validation, and testing. Prior to training, all input features were normalized using the Min-Max scaling approach~\cite{kumar2006min}, which linearly transforms each feature to the range $[-1, 1]$. This normalization improves convergence and maintains consistency throughout the training process. The transformation is defined by:

\begin{equation}
	x_i' = 2 \cdot \frac{x_i - \min(x)}{\max(x) - \min(x)} - 1
	\label{eq:minmax}
\end{equation}

This preprocessing step ensures that all features contribute proportionally during optimization, preventing dominance by features with larger numerical ranges~\cite{abdelaziz2024approximate}.

The hyperparameters employed for training the proposed architecture are listed in Table~\ref{tab:hyperparameters}. These were carefully selected to achieve a balance between model accuracy and training stability, and were tuned through empirical testing and validation performance.

\begin{figure}[!p!t]
	\centering
	\includegraphics[width=.8\columnwidth]{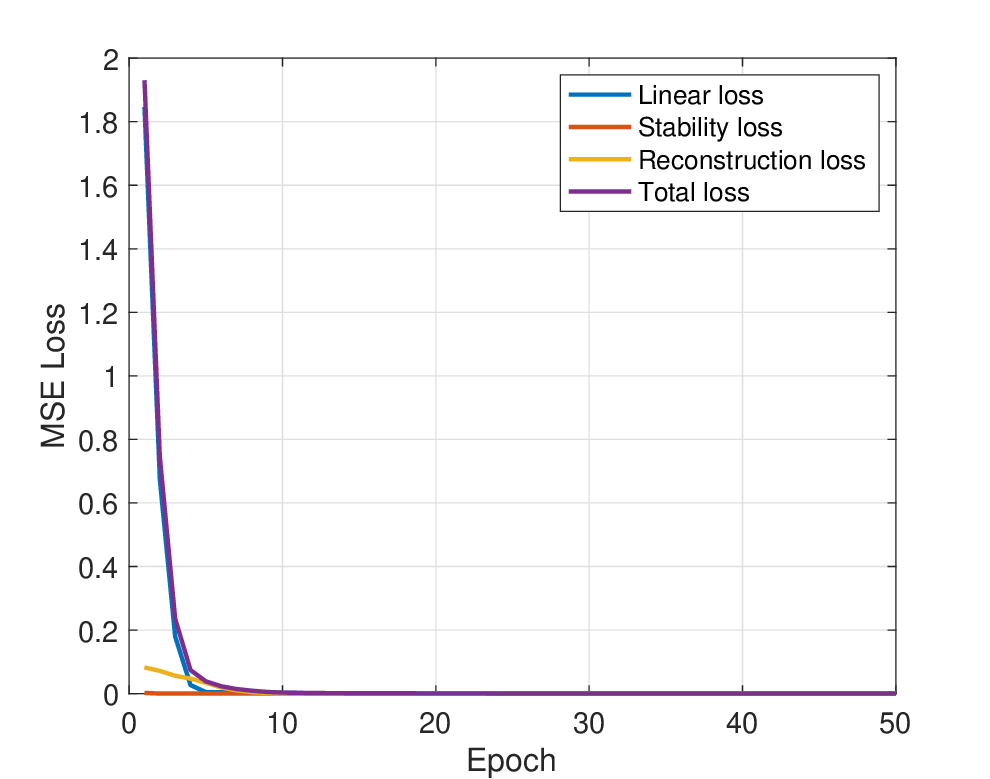}
	    \caption{Training loss convergence of the DK-MPC model over 50 epochs. The plot illustrates the Mean Squared Error (MSE) for individual loss components—linear dynamics loss, stability loss, and reconstruction loss—as well as the total loss. All components show rapid convergence within the first few epochs, indicating stable and efficient learning.}
\label{fig:loss}
\end{figure}

To evaluate the learning behavior of the proposed DK-MPC framework, we monitored the evolution of various loss components during training, as shown in Fig.~\ref{fig:loss}. The plot illustrates the convergence of the linear dynamics loss, reconstruction loss, stability loss, and the total loss over 50 epochs. All losses exhibit a rapid decline within the first 10 epochs, stabilizing at near-zero values, which indicates fast convergence and training stability. The smooth decrease in the total loss suggests that the model effectively learns a Koopman-consistent latent representation, while simultaneously satisfying reconstruction accuracy and spectral radius constraints. This behavior confirms the effectiveness of the loss design in guiding the model toward a stable and accurate linear approximation of the nonlinear system dynamics.

\begin{table}[!p!t]
	\centering
	\caption{Hyperparameters for DK-MPC Model Training}
	\label{tab:hyperparameters}
	\begin{tabular}{l c}
		\hline
		\textbf{Hyperparameter} & \textbf{Value} \\
		\hline
		Learning Rate & $1 \times 10^{-4}$ \\
		Batch Size & 32 \\
		Latent Dimension & 8 \\
		Training Epochs & 50 \\
		Optimizer & Adam \\
		Loss Coefficient $\lambda_1, \lambda_2, \lambda_3$ & 1, 50, 1\\
		Regularization Coefficient $\lambda_4$ & $1 \times 10^{-4}$ \\
		\hline
	\end{tabular}
\end{table}

\begin{figure}[!p!b]
	\centering
	\includegraphics[width=\columnwidth]{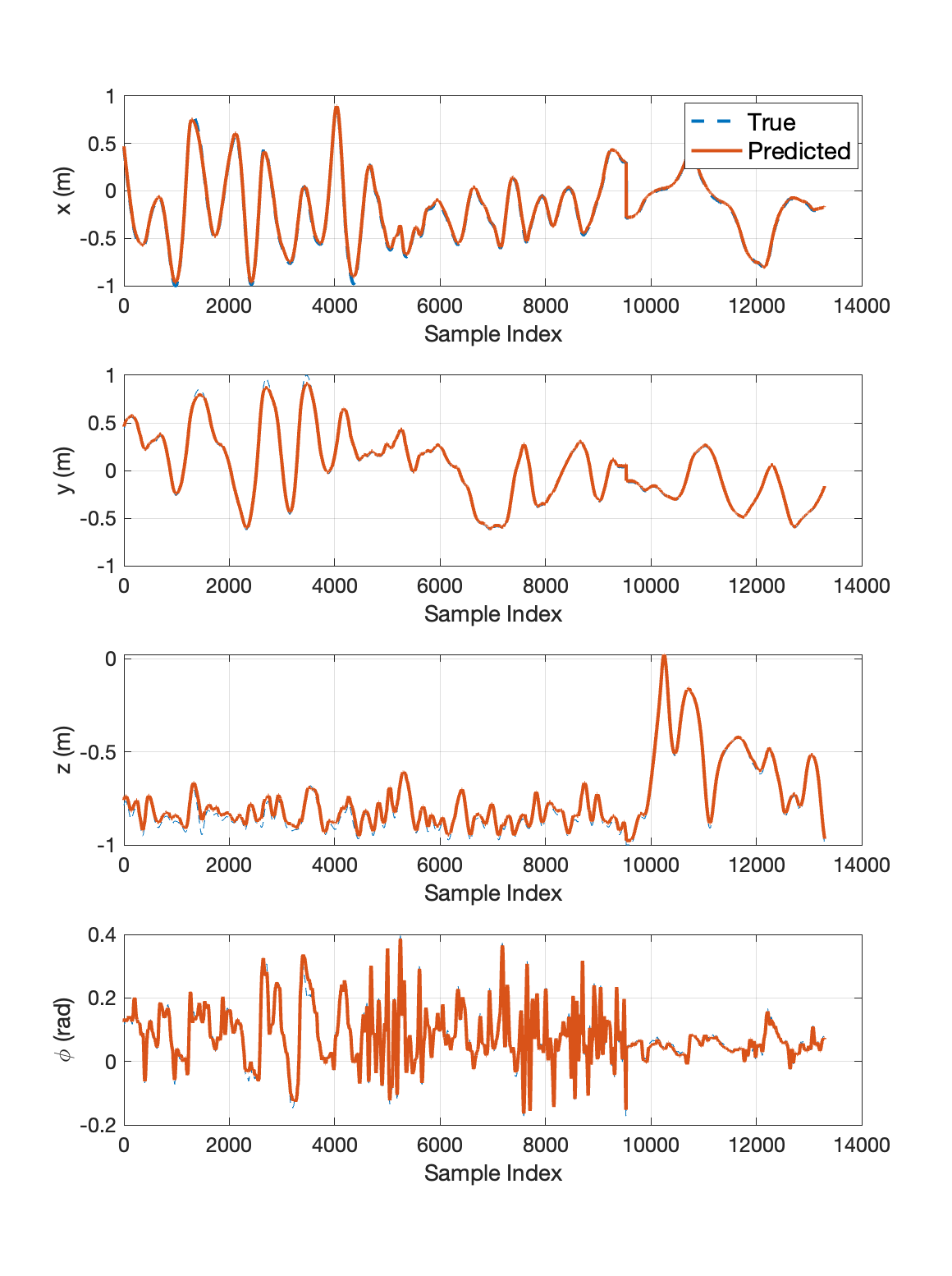}
	\caption{Prediction performance of the trained DK-MPC model. The figure shows the comparison between true and predicted trajectories for position ($x$, $y$, $z$) and roll angle ($\phi$) over time. The predicted signals (orange) closely follow the ground truth (blue dashed), demonstrating the model's ability to accurately capture the system dynamics in both translational and rotational dimensions.}
	\label{fig:comp}
\end{figure}

To evaluate the prediction accuracy of the learned Koopman-based model, we compare its outputs with the ground truth test data across multiple state dimensions, including position ($x$, $y$, $z$) and roll angle ($\phi$), as illustrated in Fig.~\ref{fig:comp}. The predicted trajectories generated by the model (shown in orange) align closely with the true states (blue dashed lines), indicating the model’s ability to capture the underlying system dynamics with high fidelity. This strong agreement across all state channels demonstrates that the lifted linear representation learned by the encoder is sufficiently expressive to approximate the nonlinear behavior of the quadrotor system. These results validate the Koopman framework’s suitability for both accurate prediction and effective control design in complex dynamical systems.

\subsection{Point Stabilization}
\begin{figure}[!p!t]
	\centering
	\includegraphics[width=.9\columnwidth]{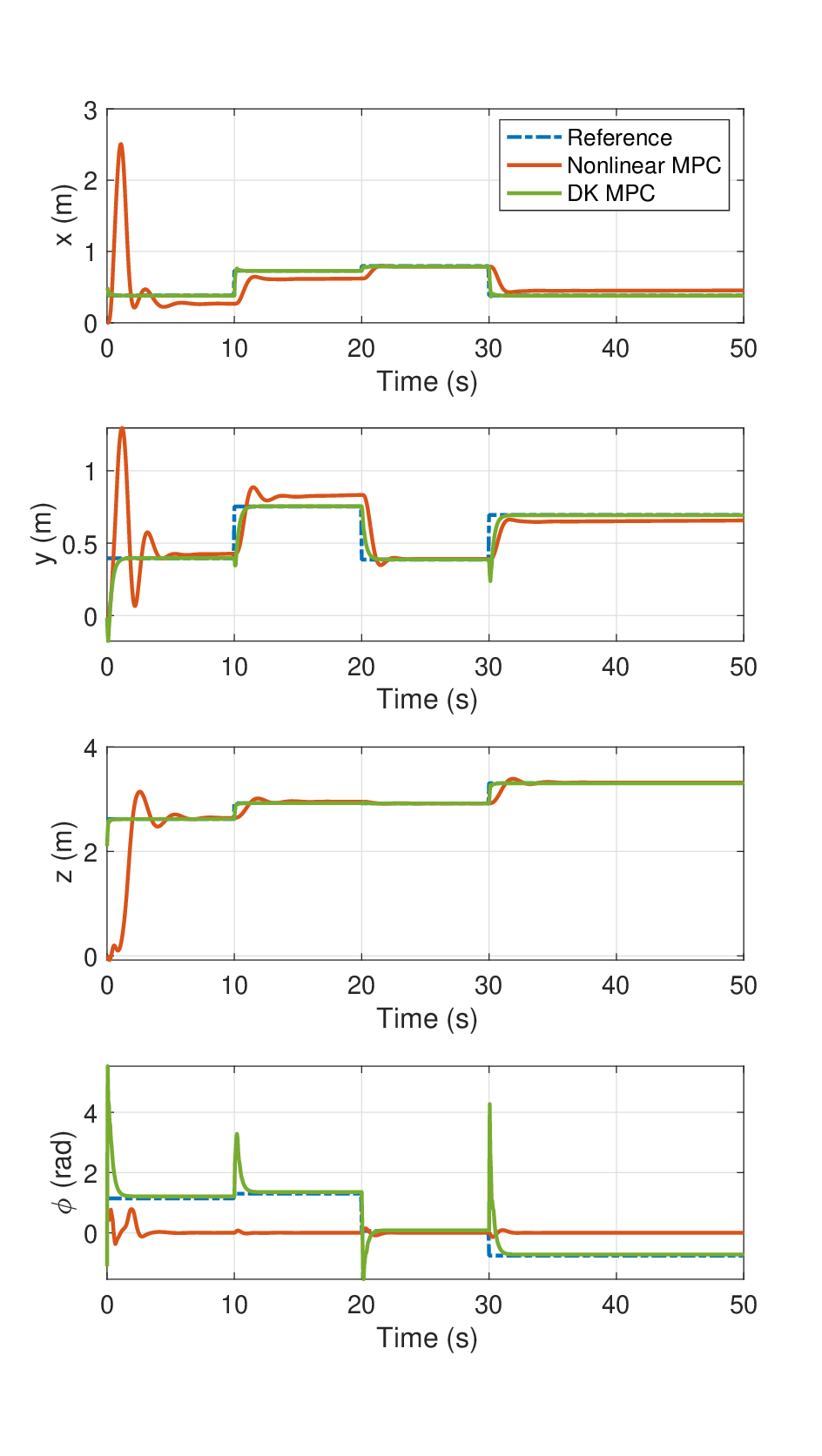}
	\caption{Point stabilization performance of the DK-MPC (green) compared to Nonlinear MPC (orange). The plots show the evolution of position ($x$, $y$, $z$) and roll angle ($\phi$) as the quadrotor tracks stepwise changes  (blue dashed) in the target position and orientation.}
	\label{fig:DK-NONLINEAR-STB}
\end{figure}

In the point stabilization experiment, the performance of DK-MPC is evaluated against a conventional Nonlinear MPC across stepwise changes in target states. As shown in Fig.~\ref{fig:DK-NONLINEAR-STB}, the DK-MPC controller demonstrates superior tracking accuracy and stability across all state dimensions. It responds quickly to abrupt changes in reference positions ($x$, $y$, $z$) and orientation ($\phi$), with minimal overshoot and short settling time. In contrast, the Nonlinear MPC exhibits noticeable oscillations, slower convergence, and greater steady-state error—particularly in the rotational dynamics. These results validate the capability of the learned Koopman-based linear representation to generalize well to unseen reference inputs, ensuring robust and efficient control in point-to-point tasks. The effectiveness of DK-MPC in handling sharp transitions also highlights its potential for real-time applications in agile and dynamic environments.

\subsection{Trajectory Tracking}
\begin{figure}[!p!b]
	\centering
	\includegraphics[width=.8\columnwidth]{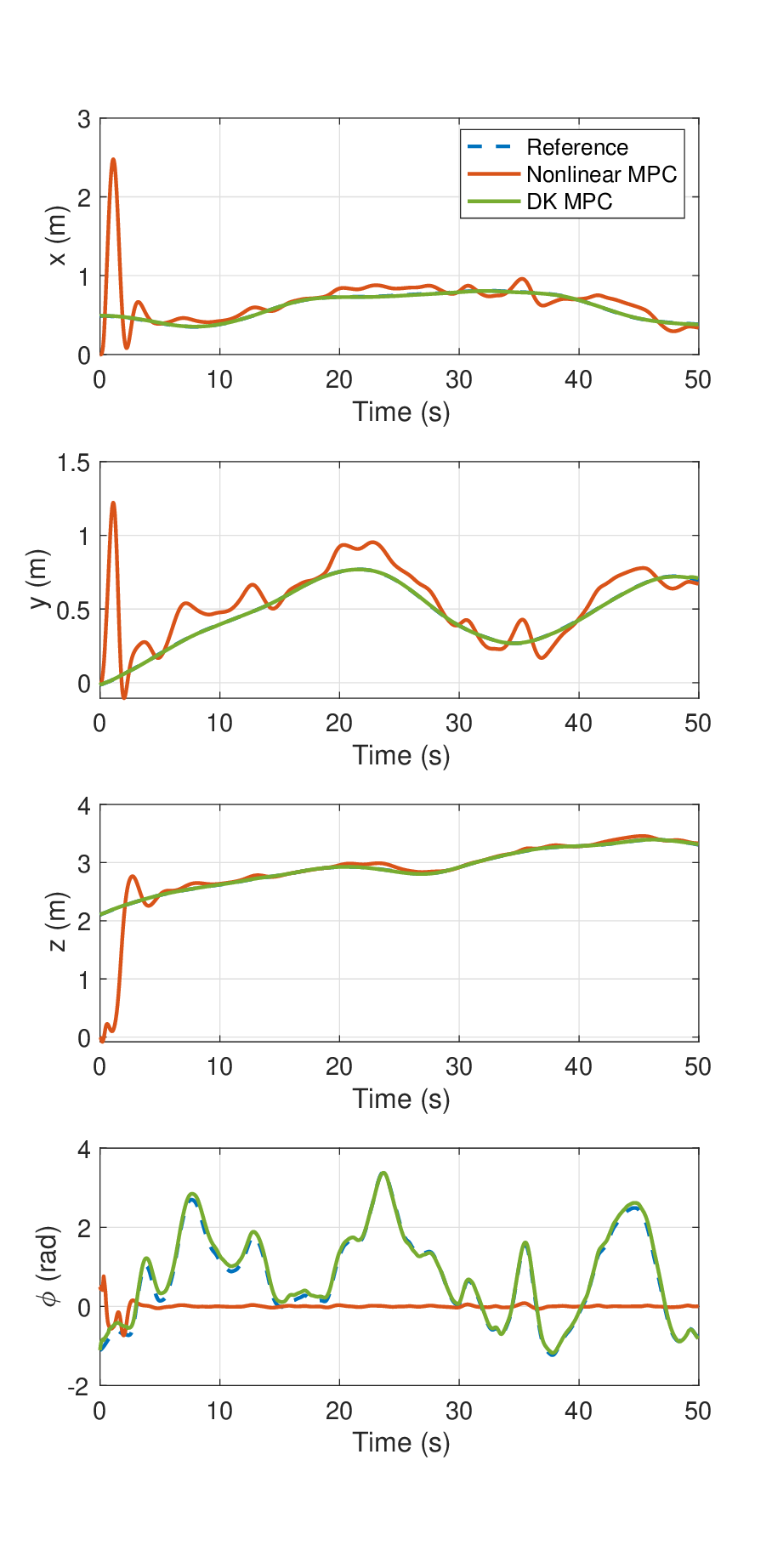}
	\caption{Comparison of control performance between the proposed DK-MPC (green) and a standard Nonlinear MPC (orange) in a trajectory tracking task (blue dashed).}
	
	\label{fig:DK-NONLINEAR}
\end{figure}

In the trajectory tracking experiment, the DK-MPC is further evaluated in a continuous motion scenario where the reference path evolves smoothly over time. As illustrated in Fig.~\ref{fig:DK-NONLINEAR}, DK-MPC consistently outperforms the baseline Nonlinear MPC across all translational ($x$, $y$, $z$) and rotational ($\phi$) states. The DK-MPC controller closely follows the reference trajectory with reduced tracking error and smoother control responses, while the Nonlinear MPC exhibits lagging behavior and higher deviation, particularly during transitions and changes in trajectory curvature. These observations emphasize the effectiveness of the Koopman-based linear model in capturing global dynamics, allowing the controller to anticipate and adjust to trajectory changes more efficiently. The results affirm that DK-MPC not only achieves high tracking accuracy but also maintains stability and robustness during dynamic flight tasks.

\subsection{Real-time Suitability}
\begin{figure}[!p!t]
	\centering
	\includegraphics[width=\columnwidth]{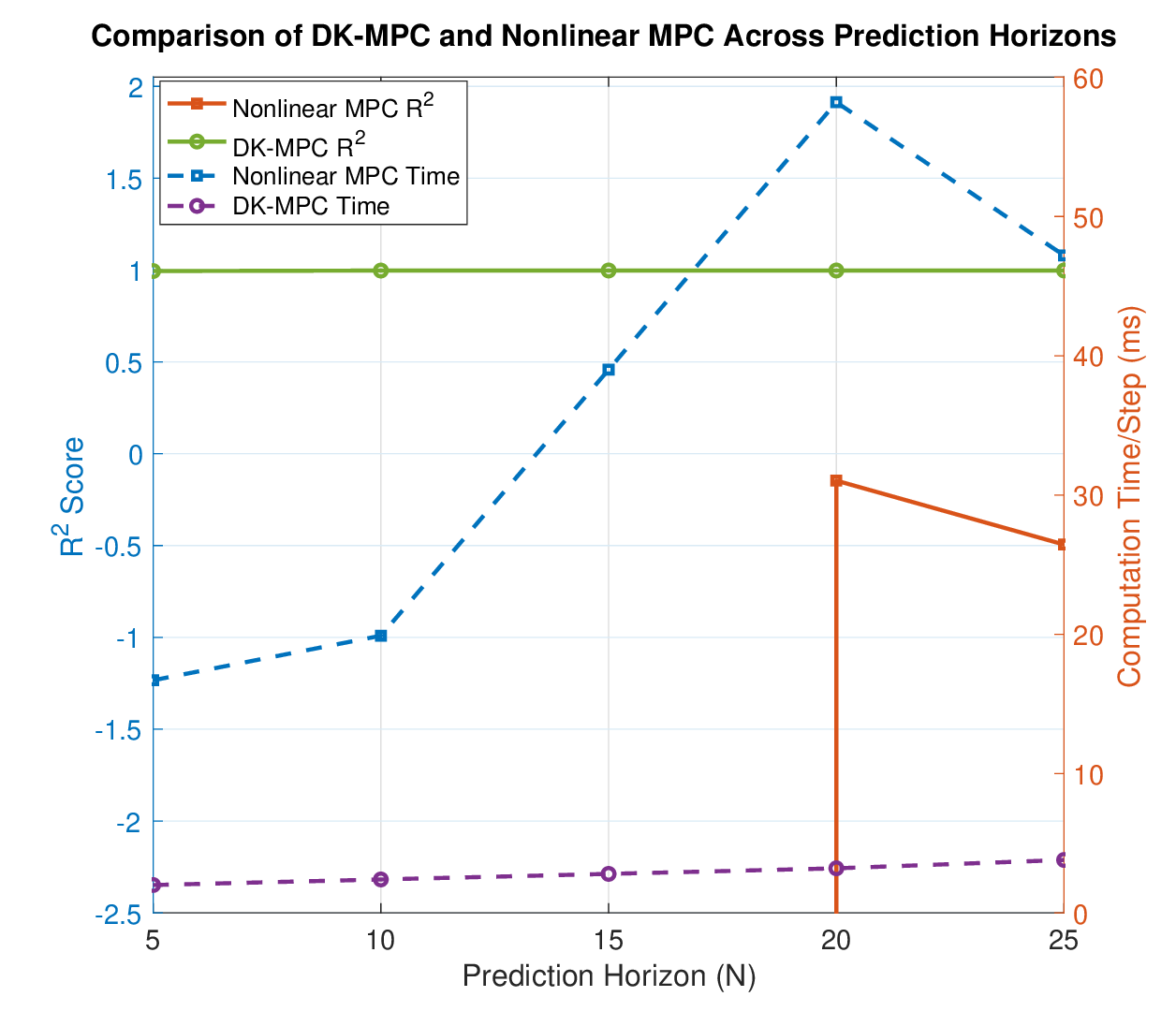}
    \caption{Comparison of DK-MPC and Nonlinear MPC performance across varying prediction horizon lengths ($N = 5$ to $25$). The left y-axis shows the $R^2$ score, indicating tracking accuracy, while the right y-axis shows the computation time per control step in milliseconds. DK-MPC maintains consistently high $R^2$ scores and low computation time across all horizons, while Nonlinear MPC exhibits significant instability in both accuracy and runtime. In the Nonlinear MPC approch where $N<20$, the $R^2$ error is significantly negative, implying that predictions are very far off from the actual values.}
	\label{fig:time-error}
\end{figure}

To further evaluate the performance and scalability of the controllers for real-time applications, we analyze how both DK-MPC and Nonlinear MPC behave across varying prediction horizon lengths. As illustrated in Fig.~\ref{fig:time-error}, DK-MPC consistently achieves high $R^2$ scores above 0.99 across all horizons, demonstrating robust prediction accuracy and reliable model generalization. In contrast, the Nonlinear MPC shows extremely poor accuracy for shorter horizons ($N = 5$, $10$, and $15$), with significantly negative $R^2$ values, indicating divergence from the reference trajectory. Only at longer horizons ($N = 20$ and $25$) does it begin to exhibit modest improvement. 

On the computational side, DK-MPC maintains a low and stable control time below 5 ms per step regardless of horizon length, highlighting its suitability for real-time applications. Nonlinear MPC, however, shows a steep increase in computational cost, reaching nearly 60 ms at $N = 20$. These results underscore the advantage of DK-MPC in achieving accurate control with minimal computational overhead, especially when scalability and fast response are critical in embedded or onboard systems.

\section{Conclusion}
\label{sec:conclude}
This study introduces a data-driven control framework for quadrotor systems, termed DK-MPC, which combines a deep Koopman operator with a model predictive controller (MPC). By utilizing sampled flight data, the deep Koopman operator learns a high-dimensional latent space where the complex, nonlinear dynamics of the quadrotor can be approximated with linear models. This linear representation enables the efficient application of MPC to compute optimal control inputs over a finite prediction horizon for accurate trajectory tracking. Numerical experimental results on the quadrotor platform demonstrate that DK-MPC achieves superior tracking performance and reduced computational cost compared to conventional nonlinear MPC approaches. Future work will focus on improving model generalization and extending the framework to more agile or disturbance-prone flight scenarios.

\bibliographystyle{plain}
\bibliography{references.bib}

\end{document}